\documentclass[final,3p,times,twocolumn]{elsarticle}
\usepackage{amssymb}
\usepackage[switch]{lineno} 

\usepackage{float}

\usepackage{balance}

\usepackage{booktabs, makecell, tabularx}
\newcolumntype{C}{>{\centering\arraybackslash}X} % centered version of "X" type

\usepackage{stfloats}
\usepackage{siunitx}

\usepackage{longtable}

\usepackage{amsmath,amssymb}
\usepackage{graphicx}

\DeclareMathOperator{\EX}{\mathbb{E}}% expected value

\usepackage{caption}
\usepackage[hyphens]{url}
\date{}

\makeatletter
\def\ps@pprintTitle{%
  \let\@oddhead\@empty
  \let\@evenhead\@empty
  \let\@oddfoot\@empty
  \let\@evenfoot\@oddfoot
}
\makeatother

\biboptions{sort&compress}

\begin{document}

\begin{frontmatter}



\title{Deep Learning for Deepfakes Creation and Detection: A Survey}


\author[inst1]{Thanh~Thi~Nguyen}
\author[inst2]{Quoc~Viet~Hung~Nguyen}
\author[inst1]{Dung~Tien~Nguyen}
\author[inst1]{Duc~Thanh~Nguyen}
\author[inst3]{Thien~Huynh-The}
\author[inst4]{Saeid~Nahavandi}
\author[inst5]{Thanh~Tam~Nguyen}
\author[inst6]{Quoc-Viet~Pham}
\author[inst7]{Cuong~M.~Nguyen}

\affiliation[inst1]{organization={School of Information Technology, Deakin University},%Department and Organization
            %addressline={Address One}, 
            %city={City One},
            %postcode={00000}, 
            state={Victoria},
            country={Australia}}

\affiliation[inst2]{organization={School of Information and Communication Technology, Griffith University},
            state={Queensland},
            country={Australia}}

\affiliation[inst3]{organization={ICT Convergence Research Center, Kumoh National Institute of Technology},
            state={Gyeongbuk},
            country={Republic of Korea}}

\affiliation[inst4]{organization={Institute for Intelligent Systems Research and Innovation, Deakin University},
            state={Victoria},
            country={Australia}}

\affiliation[inst5]{organization={Faculty of Information Technology, Ho Chi Minh City University of Technology (HUTECH)},
            state={Ho Chi Minh City},
            country={Vietnam}}

\affiliation[inst6]{organization={Korean Southeast Center for the 4th Industrial Revolution Leader Education, Pusan National University},
            state={Busan},
            country={Republic of Korea}}

\affiliation[inst7]{organization={LAMIH UMR CNRS 8201, Universite Polytechnique Hauts-de-France},
            state={Valenciennes},
            country={France}}

%\author[inst1,inst2]{Author Three}

%\affiliation[inst2]{organization={Department Two},%Department and Organization
%            addressline={Address Two}, 
%            city={City Two},
%            postcode={22222}, 
%            state={State Two},
%            country={Country Two}}

\begin{abstract}
Deep learning has been successfully applied to solve various complex problems ranging from big data analytics to computer vision and human-level control. Deep learning advances however have also been employed to create software that can cause threats to privacy, democracy and national security. One of those deep learning-powered applications recently emerged is deepfake. Deepfake algorithms can create fake images and videos that humans cannot distinguish them from authentic ones. The proposal of technologies that can automatically detect and assess the integrity of digital visual media is therefore indispensable. This paper presents a survey of algorithms used to create deepfakes and, more importantly, methods proposed to detect deepfakes in the literature to date. We present extensive discussions on challenges, research trends and directions related to deepfake technologies. By reviewing the background of deepfakes and state-of-the-art deepfake detection methods, this study provides a comprehensive overview of deepfake techniques and facilitates the development of new and more robust methods to deal with the increasingly challenging deepfakes.
\end{abstract}




\end{frontmatter}


\section{Introduction}
\label{sec1}
In a narrow definition, deepfakes (stemming from ``deep learning" and ``fake") are created by techniques that can superimpose face images of a target person onto a video of a source person to make a video of the target person doing or saying things the source person does. This constitutes a category of deepfakes, namely \emph{face-swap}. In a broader definition, deepfakes are artificial intelligence-synthesized content that can also fall into two other categories, i.e., \emph{lip-sync} and \emph{puppet-master}. Lip-sync deepfakes refer to videos that are modified to make the mouth movements consistent with an audio recording. Puppet-master deepfakes include videos of a target person (puppet) who is animated following the facial expressions, eye and head movements of another person (master) sitting in front of a camera \cite{Agarwal2019d}.

While some deepfakes can be created by traditional visual effects or computer-graphics approaches, the recent common underlying mechanism for deepfake creation is deep learning models such as autoencoders and generative adversarial networks (GANs), which have been applied widely in the computer vision domain \cite{vincent2008extracting, kingma2013auto, Goodfellow2014, makhzani2015adversarial, Tewari2018, Lin2021, Liu2021}. These models are used to examine facial expressions and movements of a person and synthesize facial images of another person making analogous expressions and movements \cite{Lyu2018web}. Deepfake methods normally require a large amount of image and video data to train models to create photo-realistic images and videos. As public figures such as celebrities and politicians may have a large number of videos and images available online, they are initial targets of deepfakes. Deepfakes were used to swap faces of celebrities or politicians to bodies in porn images and videos. The first deepfake video emerged in 2017 where face of a celebrity was swapped to the face of a porn actor. It is threatening to world security when deepfake methods can be employed to create videos of world leaders with fake speeches for falsification purposes \cite{Bloomberg2018web, Chesney2019, Hwang2020}. Deepfakes therefore can be abused to cause political or religion tensions between countries, to fool public and affect results in election campaigns, or create chaos in financial markets by creating fake news \cite{Zhou2020, Kaliyar2020, Guo2020}. It can be even used to generate fake satellite images of the Earth to contain objects that do not really exist to confuse military analysts, e.g., creating a fake bridge across a river although there is no such a bridge in reality. This can mislead a troop who have been guided to cross the bridge in a battle \cite{Tucker2019web, Fish2019web}.

As the democratization of creating realistic digital humans has positive implications, there is also positive use of deepfakes such as their applications in visual effects, digital avatars, snapchat filters, creating voices of those who have lost theirs or updating episodes of movies without reshooting them \cite{Marr2019web}. Deepfakes can have creative or productive impacts in photography, video games, virtual reality, movie productions, and entertainment, e.g., realistic video dubbing of foreign films, education through the reanimation of historical figures, virtually trying on clothes while shopping, and so on \cite{Mirsky2021, Verdoliva2020}. However, the number of malicious uses of deepfakes largely dominates that of the positive ones. The development of advanced deep neural networks and the availability of large amount of data have made the forged images and videos almost indistinguishable to humans and even to sophisticated computer algorithms. The process of creating those manipulated images and videos is also much simpler today as it needs as little as an identity photo or a short video of a target individual. Less and less effort is required to produce a stunningly convincing tempered footage. Recent advances can even create a deepfake with just a still image \cite{Zakharov2019}. Deepfakes therefore can be a threat affecting not only public figures but also ordinary people. For example, a voice deepfake was used to scam a CEO out of \$243,000 \cite{Damiani2019web}. A recent release of a software called DeepNude shows more disturbing threats as it can transform a person to a non-consensual porn \cite{Samuel2019web}. Likewise, the Chinese app Zao has gone viral lately as less-skilled users can swap their faces onto bodies of movie stars and insert themselves into well-known movies and TV clips \cite{Guardian2019web}. These forms of falsification create a huge threat to violation of privacy and identity, and affect many aspects of human lives.

\begin{figure}[ht]
\centering
\includegraphics[width=0.85\columnwidth]{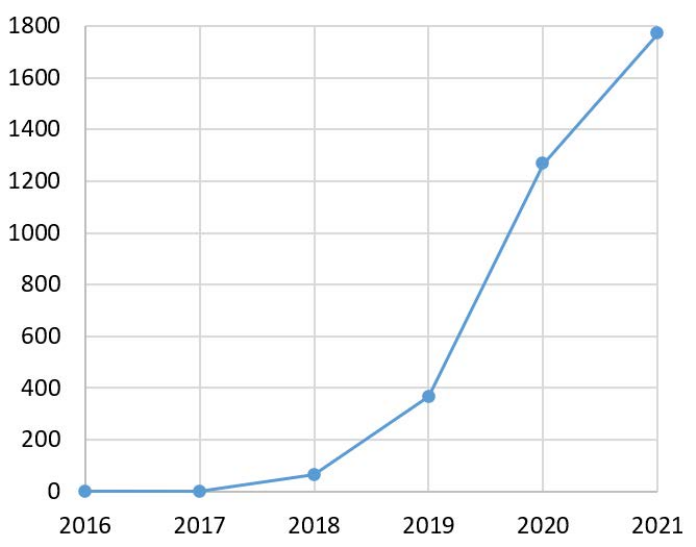}
\caption{Number of papers related to deepfakes in years from 2016 to 2021, obtained from https://app.dimensions.ai at the end of 2021 with the search keyword ``deepfake" applied to full text of scholarly papers.}
\label{fig0} 
\end{figure}

Finding the truth in digital domain therefore has become increasingly critical. It is even more challenging when dealing with deepfakes as they are majorly used to serve malicious purposes and almost anyone can create deepfakes these days using existing deepfake tools. Thus far, there have been numerous methods proposed to detect deepfakes \cite{Lyu2020, Guarnera2020e, Jafar2020, Trinh2020, Younus2020}. Most of them are based on deep learning, and thus a battle between malicious and positive uses of deep learning methods has been arising. To address the threat of face-swapping technology or deepfakes, the United States Defense Advanced Research Projects Agency (DARPA) initiated a research scheme in media forensics (named Media Forensics or MediFor) to accelerate the development of fake digital visual media detection methods \cite{Turek2019web}. Recently, Facebook Inc. teaming up with Microsoft Corp and the Partnership on AI coalition have launched the Deepfake Detection Challenge to catalyse more research and development in detecting and preventing deepfakes from being used to mislead viewers \cite{Schroepfer2019web}. Data obtained from https://app.dimensions.ai at the end of 2021 show that the number of deepfake papers has increased significantly in recent years (Fig. \ref{fig0}). Although the obtained numbers of deepfake papers may be lower than actual numbers but the research trend of this topic is obviously increasing.

\begin{figure}[!t]
\centering
\includegraphics[width=1.0\columnwidth]{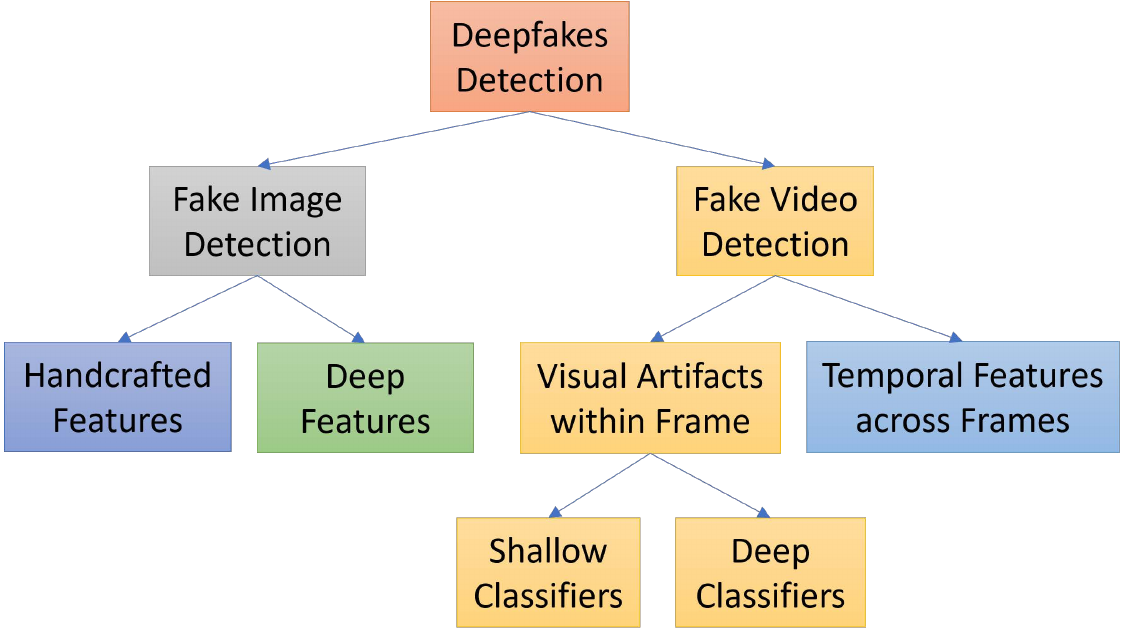}
\caption{Categories of reviewed papers relevant to deepfake detection methods where we divide papers into two major groups, i.e., fake image detection and face video detection.}
\label{fig2} 
\end{figure}

There have been existing survey papers about creating and detecting deepfakes, presented in \cite{Tolosana2020, Verdoliva2020, Mirsky2021}. For example, \citet{Mirsky2021} focused on reenactment approaches (i.e., to change a target’s expression, mouth, pose, gaze or body), and replacement approaches (i.e., to replace a target’s face by swap or transfer methods). \citet{Verdoliva2020} separated detection approaches into conventional methods (e.g., blind methods without using any external data for training, one-class sensor-based and model-based methods, and supervised methods with handcrafted features) and deep learning-based approaches (e.g., CNN models). \citet{Tolosana2020} categorized both creation and detection methods based on the way deepfakes are created, including entire face synthesis, identity swap, attribute manipulation, and expression swap. On the other hand, we carry out the survey with a different perspective and taxonomy. We categorize the deepfake detection methods based on the data type, i.e., images or videos, as presented in Fig. \ref{fig2}. With fake image detection methods, we focus on the features that are used, i.e., whether they are handcrafted features or deep features. With fake video detection methods, two main subcategories are identified based on whether the method uses temporal features across frames or visual artifacts within a video frame. We also discuss extensively the challenges, research trends and directions on deepfake detection and multimedia forensics problems.

\section{Deepfake Creation}
\label{sec2}

\begin{figure}[t]
\centering
\includegraphics[width=0.92\columnwidth]{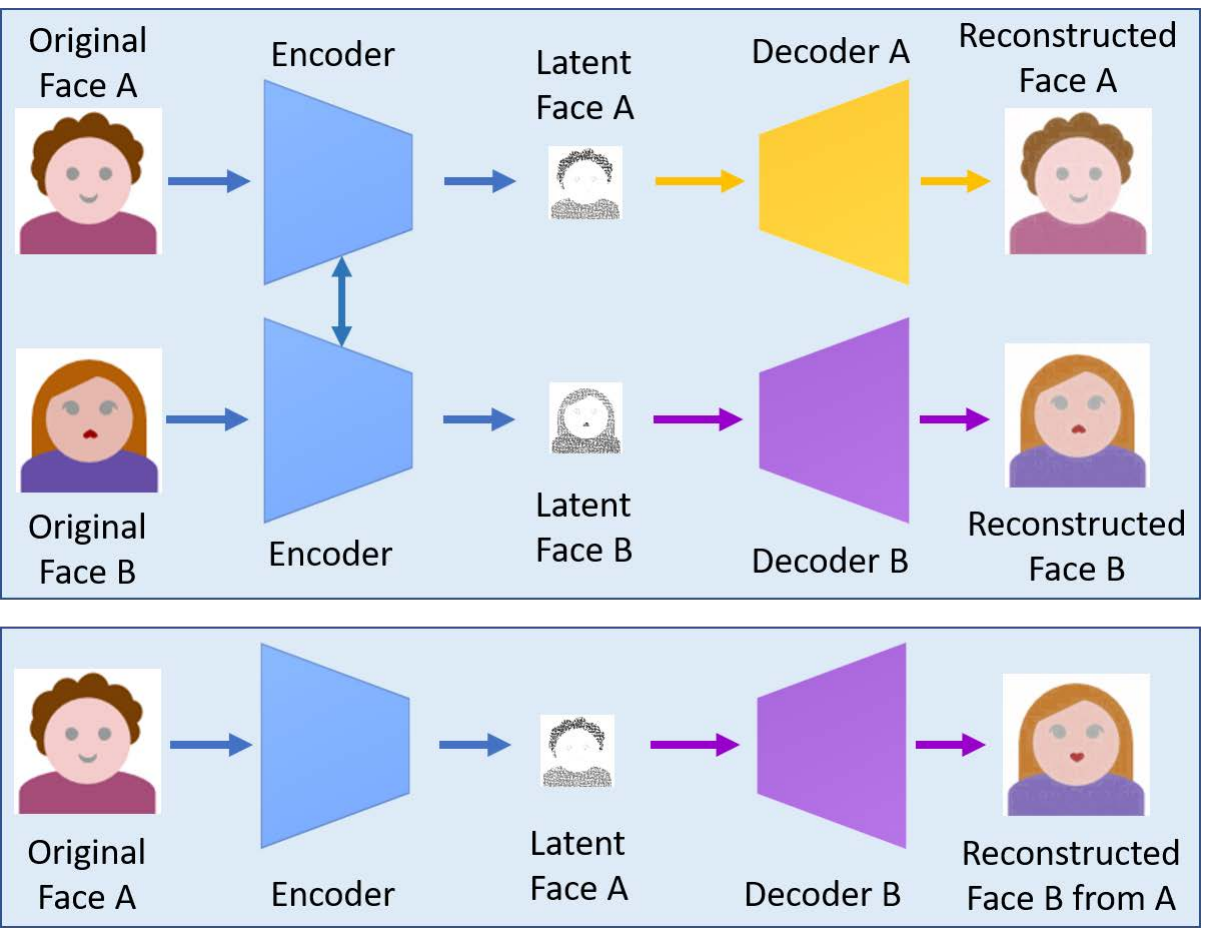}
\caption{A deepfake creation model using two encoder-decoder pairs. Two networks use the same encoder but different decoders for training process (top). An image of face A is encoded with the common encoder and decoded with decoder B to create a deepfake (bottom). The reconstructed image (in the bottom) is the face B with the mouth shape of face A. Face B originally has the mouth of an upside-down heart while the reconstructed face B has the mouth of a conventional heart.}
\label{fig1} 
\end{figure}

Deepfakes have become popular due to the quality of tampered videos and also the easy-to-use ability of their applications to a wide range of users with various computer skills from professional to novice. These applications are mostly developed based on deep learning techniques. Deep learning is well known for its capability of representing complex and high-dimensional data. One variant of the deep networks with that capability is deep autoencoders, which have been widely applied for dimensionality reduction and image compression \cite{Punnappurath2019, Cheng2019, Chorowski2019}. The first attempt of deepfake creation was FakeApp, developed by a Reddit user using autoencoder-decoder pairing structure \cite{Githubweb, FakeAppweb}. In that method, the autoencoder extracts latent features of face images and the decoder is used to reconstruct the face images. To swap faces between source images and target images, there is a need of two encoder-decoder pairs where each pair is used to train on an image set, and the encoder's parameters are shared between two network pairs. In other words, two pairs have the same encoder network. This strategy enables the common encoder to find and learn the similarity between two sets of face images, which are relatively unchallenging because faces normally have similar features such as eyes, nose, mouth positions. Fig. \ref{fig1} shows a deepfake creation process where the feature set of face A is connected with the decoder B to reconstruct face B from the original face A. This approach is applied in several works such as DeepFaceLab \cite{DeepFaceLabweb}, DFaker \cite{DFakerweb}, DeepFake\_tf (tensorflow-based deepfakes) \cite{DeepFake_tfweb}.

\begin{table*}
\centering
\begin{scriptsize}
\caption{Summary of notable deepfake tools}
\label{table1}
\begin{tabular}{p{0.10\textwidth} p{0.29\textwidth} p{0.54\textwidth}}
\hline
\textbf{Tools} & \textbf{Links} & \textbf{Key Features}\\
\hline
Faceswap
&https://github.com/deepfakes/faceswap
&- Using two encoder-decoder pairs.\newline
- Parameters of the encoder are shared.\\
\hline
Faceswap-GAN
&https://github.com/shaoanlu/faceswap-GAN
&Adversarial loss and perceptual loss (VGGface) are added to an auto-encoder architecture.\\
\hline
Few-Shot Face Translation
& https://github.com/shaoanlu/fewshot-face-translation-GAN
& - Use a pre-trained face recognition model to extract latent embeddings for GAN processing.\newline
- Incorporate semantic priors obtained by modules from FUNIT \cite{Liu2019e} and SPADE \cite{Park2019e}.\\
\hline
DeepFaceLab
&https://github.com/iperov/DeepFaceLab
&- Expand from the Faceswap method with new models, e.g. H64, H128, LIAEF128, SAE \cite{mrdfweb}.\newline
- Support multiple face extraction modes, e.g. S3FD, MTCNN, dlib, or manual \cite{mrdfweb}.\\
\hline
DFaker
&https://github.com/dfaker/df
&- DSSIM loss function \cite{DSSIMweb} is used to reconstruct face.\newline
- Implemented based on Keras library.\\
\hline
DeepFake\_tf
&https://github.com/StromWine/DeepFake\_tf
&Similar to DFaker but implemented based on tensorflow.\\
\hline
%Deepfakes web $\beta$
%& https://deepfakesweb.com/
%& Commercial website for face swapping using deep learning algorithms.\\
% \hline
AvatarMe
& https://github.com/lattas/AvatarMe
& -	Reconstruct 3D faces from arbitrary ``in-the-wild" images.\newline
- Can reconstruct authentic 4K by 6K-resolution 3D faces from a single low-resolution image \cite{Lattas2020}.\\
\hline
MarioNETte
& https://hyperconnect.github.io/MarioNETte
& -	A few-shot face reenactment framework that preserves the target identity.\newline
- No additional fine-tuning phase is needed for identity adaptation \cite{Ha2020}.\\
\hline
DiscoFaceGAN
& https://github.com/microsoft/DiscoFaceGAN
& -	Generate face images of virtual people with independent latent variables of identity, expression, pose, and illumination.\newline
- Embed 3D priors into adversarial learning \cite{Deng2020}.\\
\hline
StyleRig
& https://gvv.mpi-inf.mpg.de/projects/StyleRig
& -	Create portrait images of faces with a rig-like control over a pretrained and fixed StyleGAN via 3D morphable face models.\newline
- Self-supervised without manual annotations \cite{Tewari2020}.\\
\hline
FaceShifter
& https://lingzhili.com/FaceShifterPage
& -	Face swapping in high-fidelity by exploiting and integrating the target attributes.\newline
- Can be applied to any new face pairs without requiring subject specific training \cite{Li2019e}.\\
\hline
FSGAN
& https://github.com/YuvalNirkin/fsgan
& -	A face swapping and reenactment model that can be applied to pairs of faces without requiring training on those faces.\newline
- Adjust to both pose and expression variations \cite{Nirkin2019}.\\
\hline
StyleGAN
& https://github.com/NVlabs/stylegan
& -	A new generator architecture for GANs is proposed based on style transfer literature.\newline
- The new architecture leads to automatic, unsupervised separation of high-level attributes and enables intuitive, scale-specific control of the synthesis of images \cite{Karras2019}.\\
\hline
Face2Face & https://justusthies.github.io/posts/face2face/ & - Real-time facial reenactment of monocular target video sequence, e.g. Youtube video.\newline
- Animate the facial expressions of the target video by a source actor and re-render the manipulated output video in a photo-realistic fashion \cite{Thies2016}.\\
\hline
Neural Textures & https://github.com/SSRSGJYD/NeuralTexture & 
- Feature maps that are learned as part of the scene capture process and stored as maps on top of {3D} mesh proxies.\newline
- Can coherently re-render or manipulate existing video content in both static and dynamic environments at real-time rates \cite{thies2019deferred}.\\
\hline
Transformable Bottleneck Networks
& https://github.com/kyleolsz/TB-Networks
& -	A method for fine-grained 3D manipulation of image content.\newline
- Apply spatial transformations in CNN models using a transformable bottleneck framework \cite{Olszewski2019}.\\
\hline
``Do as I Do" Motion \newline Transfer
& github.com/carolineec/EverybodyDanceNow
& - Automatically transfer the motion from a source to a target person by learning a video-to-video translation.\newline
-  Can create a motion-synchronized dancing video with multiple subjects \cite{Chan2019}.\\
\hline
Neural Voice Puppetry
& https://justusthies.github.io/posts/neural-voice-puppetry
& - A method for audio-driven facial video synthesis.\newline
- Synthesize videos of a talking head from an audio sequence of another person using 3D face representation. \cite{Thies2020}.\\
\hline
\end{tabular}
\end{scriptsize}
\end{table*}

By adding adversarial loss and perceptual loss implemented in VGGFace \cite{VGGFaceweb} to the encoder-decoder architecture, an improved version of deepfakes based on the generative adversarial network \cite{Goodfellow2014}, i.e., faceswap-GAN, was proposed in \cite{Faceswap-GANweb}. The VGGFace perceptual loss is added to make eye movements to be more realistic and consistent with input faces and help to smooth out artifacts in segmentation mask, leading to higher quality output videos. This model facilitates the creation of outputs with 64x64, 128x128, and 256x256 resolutions. In addition, the multi-task convolutional neural network (CNN) from the FaceNet implementation \cite{FaceNetweb} is used to make face detection more stable and face alignment more reliable. The CycleGAN \cite{CycleGANweb} is utilized for generative network implementation in this model.

\begin{figure}[!ht]
\centering
\includegraphics[width=0.55\columnwidth]{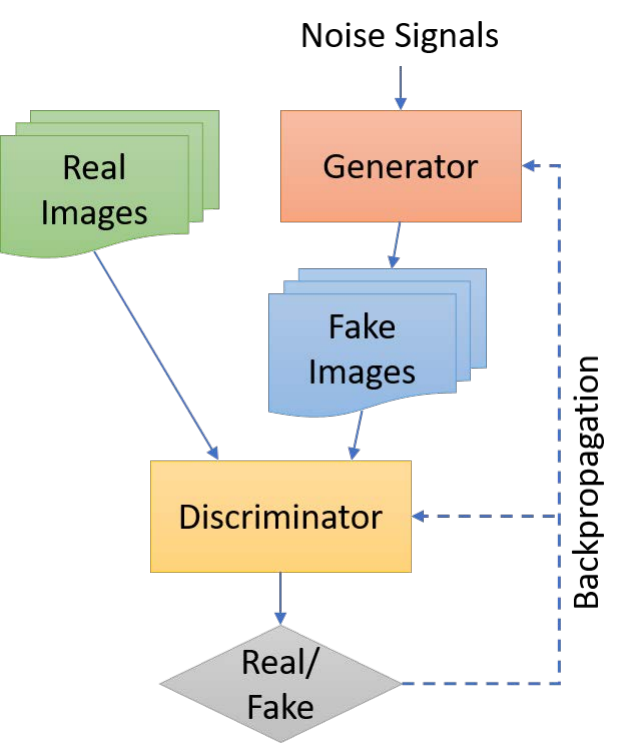}
\caption{The GAN architecture consisting of a generator and a discriminator, and each can be implemented by a neural network. The entire system can be trained with backpropagation that allows both networks to improve their capabilities.}
\label{figGAN} 
\end{figure}

A conventional GAN model comprises two neural networks: a generator and a discriminator as depicted in Fig. \ref{figGAN}. Given a dataset of real images $x$ having a distribution of $p_{data}$, the aim of the generator $G$ is to produce images $G(z)$ similar to real images $x$ with $z$ being noise signals having a distribution of $p_z$. The aim of the discriminator $G$ is to correctly classify images generated by $G$ and real images $x$. The discriminator $D$ is trained to improve its classification capability, i.e., to maximize $D(x)$, which represents the probability that $x$ is a real image rather than a fake image generated by $G$. On the other hand, $G$ is trained to minimize the probability that its outputs are classified by $D$ as synthetic images, i.e., to minimize $1-D(G(z))$. This is a minimax game between two players $D$ and $G$ that can be described by the following value function \cite{Goodfellow2014}:
\begin{multline}
    \min_G \max_D V(D,G)=\EX_{x\sim p_{data}(x)}[\log D(x)] \\ + \EX_{z\sim p_z(z)}[\log (1-D(G(z)))]
\end{multline}
After sufficient training, both networks improve their capabilities, i.e., the generator $G$ is able to produce images that are really similar to real images while the discriminator $D$ is highly capable of distinguishing fake images from real ones.

%\begin{figure}[!htp]
%\centering
%\includegraphics[width=1.0\columnwidth]{styleGAN.PNG}
%\caption{A structure comparison between two generators: one of a PGGAN \cite{Karras2017} (a) and another of a StyleGAN \cite{Karras2019} (b). In PGGAN, the latent code is fed to the input layer only. In StyleGAN, the latent code is first mapped into an intermediate latent space $W$, which is then injected into the generator via the adaptive instance normalization (AdaIN) at each convolution layer. Gaussian noise is added after each convolution, but before the AdaIN operations \cite{Karras2019}.}
%\label{figStyleGAN} 
%\end{figure}

\begin{figure}[h]
\centering
\includegraphics[width=0.9\columnwidth]{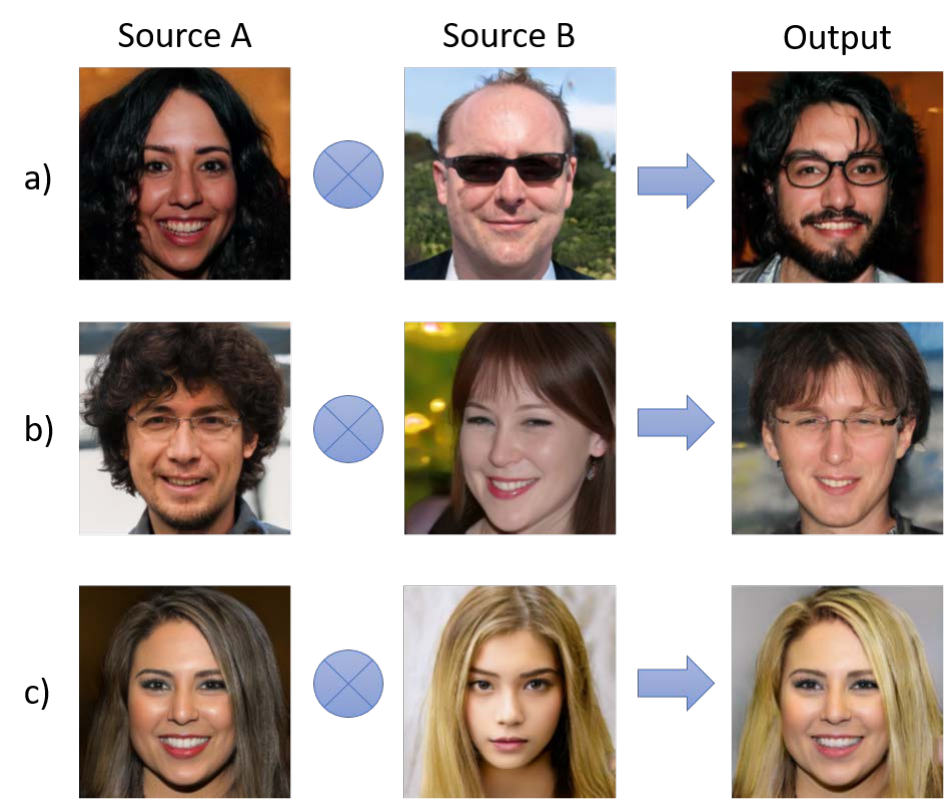}
\caption{Examples of mixing styles using StyleGAN: the output images are generated by copying a specified subset of styles from source B and taking the rest from source A. a) Copying coarse styles from source B will generate images that have high-level aspects from source B and all colors and finer facial features from source A; b) if copying the styles of middle resolutions from B, the output images will have smaller scale facial features from B and preserve the pose, general face shape, and eyeglasses from A; c) if copying the fine styles from source B, the generated images will have the color scheme and microstructure of source B \cite{Karras2019}.}
\label{figMixing} 
\end{figure}

Table \ref{table1} presents a summary of popular deepfake tools and their typical features. Among them, a prominent method for face synthesis based on a GAN model, namely StyleGAN, was introduced in \cite{Karras2019}. StyleGAN is motivated by style transfer \cite{Huang2017} with a special generator network architecture that is able to create realistic face images. In a traditional GAN model, e.g., the progressive growing of GAN (PGGAN) \cite{Karras2017}, the signal noise (latent code) is fed to the input layer of a feedforward network that represents the generator. In StyleGAN, there are two networks constructed and linked together, a mapping network $f$ and a synthesis network $g$. The latent code $z \in Z$ is first converted to $w \in W$ (where $W$ is an intermediate latent space) through a non-linear function $f:Z\rightarrow W$, which is characterized by a neural network (i.e., the mapping network) consisting of several fully connected layers. Using an affine tranformation, the intermediate representation $w$ is specialized to styles $y=(y_s,y_b)$ that will be fed to the adaptive instance normalization (AdaIN) operations, specified as: 
\begin{equation}
    \mathrm{AdaIN}(x_i,y)=y_{s,i}\frac{x_i-\mu(x_i)}{\sigma(x_i)}+y_{b,i}
\end{equation}
where each feature map $x_i$ is normalized separately. The StyleGAN generator architecture allows controlling the image synthesis by modifying the styles via different scales. In addition, instead of using one random latent code during training, this method uses two latent codes to generate a given proportion of images. More specifically, two latent codes $z_1$ and $z_2$ are fed to the mapping network to create respectively $w_1$ and $w_2$ that control the styles by applying $w_1$ before and $w_2$ after the crossover point. Fig. \ref{figMixing} demonstrates examples of images created by mixing two latent codes at three different scales where each subset of styles controls separate meaningful high-level attributes of the image. In other words, the generator architecture of StyleGAN is able to learn separation of high-level attributes (e.g., pose and identity when trained on human faces) and enables intuitive, scale-speciﬁc control of the face synthesis.

\section{Deepfake Detection}
\label{sec:3}
Deepfake detection is normally deemed a binary classification problem where classifiers are used to classify between authentic videos and tampered ones. This kind of methods requires a large database of real and fake videos to train classification models. The number of fake videos is increasingly available, but it is still limited in terms of setting a benchmark for validating various detection methods. To address this issue, Korshunov and Marcel \cite{Korshunov2019} produced a notable deepfake dataset consisting of 620 videos based on the GAN model using the open source code Faceswap-GAN \cite{Faceswap-GANweb}. Videos from the publicly available VidTIMIT database \cite{vidtimit} were used to generate low and high quality deepfake videos, which can effectively mimic the facial expressions, mouth movements, and eye blinking. These videos were then used to test various deepfake detection methods. Test results show that the popular face recognition systems based on VGG \cite{Parkhi2015} and Facenet \cite{FaceNetweb, Schroff2015} are unable to detect deepfakes effectively. Other methods such as lip-syncing approaches \cite{Chung2017, Suwajanakorn2017, Korshunov2018} and image quality metrics with support vector machine (SVM) \cite{Galbally2014} produce very high error rate when applied to detect deepfake videos from this newly produced dataset. This raises concerns about the critical need of future development of more robust methods that can detect deepfakes from genuine.

This section presents a survey of deepfake detection methods where we group them into two major categories: fake image detection methods and fake video detection ones (Fig. \ref{fig2}). The latter is distinguished into two smaller groups: \emph{visual artifacts} within single video frame-based methods and \emph{temporal features} across frames-based ones. Whilst most of the methods based on temporal features use deep learning \emph{recurrent} classification models, the methods use visual artifacts within video frame can be implemented by either deep or shallow classifiers.

\subsection{Fake Image Detection}
Deepfakes are increasingly detrimental to privacy, society security and democracy \cite{Chesney2018}. Methods for detecting deepfakes have been proposed as soon as this threat was introduced. Early attempts were based on handcrafted features obtained from artifacts and inconsistencies of the fake image synthesis process. Recent methods, e.g., \cite{deLima2020, Amerini2020}, have commonly applied deep learning to automatically extract salient and discriminative features to detect deepfakes.

\subsubsection{Handcrafted Features-based Methods}
Most works on detection of GAN generated images do not consider the generalization capability of the detection models although the development of GAN is ongoing, and many new extensions of GAN are frequently introduced. Xuan et al. \cite{Xuan2019} used an image preprocessing step, e.g., Gaussian blur and Gaussian noise, to remove low level high frequency clues of GAN images. This increases the pixel level statistical similarity between real images and fake images and allows the forensic classifier to learn more intrinsic and meaningful features, which has better generalization capability than previous image forensics methods \cite{Yang2016, Bayar2016} or image steganalysis networks \cite{Qian2015}.

Zhang et al. \cite{Zhang2017} used the bag of words method to extract a set of compact features and fed it into various classifiers such as SVM \cite{Wang2017}, random forest (RF) \cite{Bai2017} and multi-layer perceptrons (MLP) \cite{Zheng2016} for discriminating swapped face images from the genuine. Among deep learning-generated images, those synthesised by GAN models are probably most difficult to detect as they are realistic and high-quality based on GAN's capability to learn distribution of the complex input data and generate new outputs with similar input distribution. 

On the other hand, Agarwal and Varshney \cite{Agarwal2019} cast the GAN-based deepfake detection as a hypothesis testing problem where a statistical framework was introduced using the information-theoretic study of authentication \cite{Maurer2000}. The minimum distance between distributions of legitimate images and images generated by a particular GAN is defined, namely the oracle error. The analytic results show that this distance increases when the GAN is less accurate, and in this case, it is easier to detect deepfakes. In case of high-resolution image inputs, an extremely accurate GAN is required to generate fake images that are hard to detect by this method.

\subsubsection{Deep Features-based Methods}
Face swapping has a number of compelling applications in video compositing, transfiguration in portraits, and especially in identity protection as it can replace faces in photographs by ones from a collection of stock images. However, it is also one of the techniques that cyber attackers employ to penetrate identification or authentication systems to gain illegitimate access. The use of deep learning such as CNN and GAN has made swapped face images more challenging for forensics models as it can preserve pose, facial expression and lighting of the photographs \cite{Korshunova2017}.

Hsu et al. \cite{Hsu2019} introduced a two-phase deep learning method for detection of deepfake images. The first phase is a feature extractor based on the common fake feature network (CFFN) where the Siamese network architecture presented in \cite{Chopra2005} is used. The CFFN encompasses several dense units with each unit including different numbers of dense blocks \cite{Huang2017} to improve the representative capability for the input images. 
%The number of dense units is three or five depending on the validation data being face or general images, and the number of channels in each unit is varied up to a few hundreds.
Discriminative features between the fake and real images are extracted through the CFFN learning process based on the use of pairwise information, which is the label of each pair of two input images. If the two images are of the same type, i.e., fake-fake or real-real, the pairwise label is $1$. In contrast, if they are of different types, i.e., fake-real, the pairwise label is $0$. The CFFN-based discriminative features are then fed to a neural network classifier to distinguish deceptive images from genuine. The proposed method is validated for both fake face and fake general image detection. On the one hand, the face dataset is obtained from CelebA \cite{Liu2015}, containing 10,177 identities and 202,599 aligned face images of various poses and background clutter. Five GAN variants are used to generate fake images with size of 64x64, including deep convolutional GAN (DCGAN) \cite{Radford2015}, Wasserstein GAN (WGAN) \cite{Arjovsky2017}, WGAN with gradient penalty (WGAN-GP) \cite{Gulrajani2017}, least squares GAN \cite{Mao2017}, and PGGAN \cite{Karras2017}. A total of 385,198 training images and 10,000 test images of both real and fake ones are obtained for validating the proposed method. On the other hand, the general dataset is extracted from the ILSVRC12 \cite{Russakovsky2015}. The large scale GAN training model for high fidelity natural image synthesis (BIGGAN) \cite{Brock2018}, self-attention GAN \cite{Zhang2018} and spectral normalization GAN \cite{Miyato2018} are used to generate fake images with size of 128x128. The training set consists of 600,000 fake and real images whilst the test set includes 10,000 images of both types. Experimental results show the superior performance of the proposed method against its competing methods such as those introduced in \cite{Farid2009, Mo2018, Marra2018, Hsu2018}.

Likewise, \citet{guo2021blind} proposed a CNN model, namely SCnet, to detect deepfake images, which are generated by the Glow-based facial forgery tool \cite{kingma2018glow}. The fake images synthesized by the Glow model \cite{kingma2018glow} have the facial expression maliciously tampered. These images are hyper-realistic with perfect visual qualities, but they still have subtle or noticeable manipulation traces, which are exploited by the SCnet. The SCnet is able to automatically learn high-level forensics features of image data thanks to a hierarchical feature extraction block, which is formed by stacking four convolutional layers. Each layer learns a new set of feature maps from the previous layer, with each convolutional operation is defined by:
\begin{equation}
    f_j^{(n)} = \sum_{i=1}^i f_i^{(n-1)}*\omega_{ij}^{(n)} + b_j^{(n)}
\end{equation}
where $f_j^{(n)}$ is the $j^{th}$ feature map of the $n^{th}$ layer, $\omega_{ij}^{(n)}$ is the weight of the $i^{th}$ channel of the $j^{th}$ convolutional kernel in the $n^{th}$ layer, and $b_j^{(n)}$ is the bias term of the $j^{th}$ convolutional kernel in the $n^{th}$ layer.
The proposed approach is evaluated using a dataset consisting of 321,378 face images, which are created by applying the Glow model \cite{kingma2018glow} to the CelebA face image dataset \cite{Liu2015}. Evaluation results show that the SCnet model obtains higher accuracy and better generalization than the Meso-4 model proposed in \cite{Afchar2018}.

\begin{figure*}[!ht]
\centering
\includegraphics[width=1.75\columnwidth]{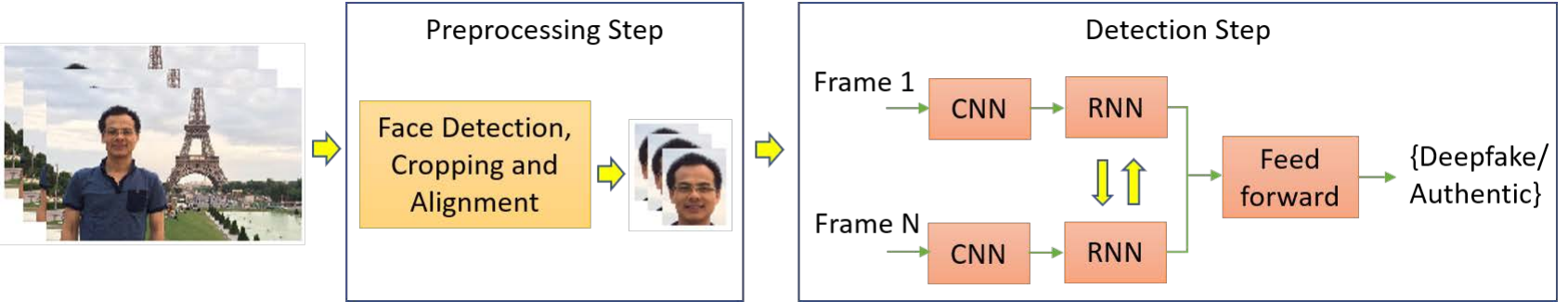}
\caption{A two-step process for face manipulation detection where the preprocessing step aims to detect, crop and align faces on a sequence of frames and the second step distinguishes manipulated and authentic face images by combining convolutional neural network (CNN) and recurrent neural network (RNN) \cite{Sabir2019}.}
\label{fig3} 
\end{figure*}

Recently, \citet{zhao2021learning} proposed a method for deepfake detection using self-consistency of local source features, which are content-independent, spatially-local information of images. These features could come from either imaging pipelines, encoding methods or image synthesis approaches. The hypothesis is that a modified image would have different source features at different locations, while an original image will have the same source features across locations. These source features, represented in the form of down-sampled feature maps, are extracted by a CNN model using a special representation learning method called pairwise self-consistency learning. This learning method aims to penalize pairs of feature vectors that refer to locations from the same image for having a low cosine similarity score. At the same time, it also penalizes the pairs from different images for having a high similarity score. The learned feature maps are then fed to a classification method for deepfake detection. This proposed approach is evaluated on seven popular datasets, including FaceForensics++ \cite{Rossler2019}, DeepfakeDetection \cite{Dufour2019}, Celeb-DF-v1 \& Celeb-DF-v2 \cite{Li2020}, Deepfake Detection Challenge (DFDC) \cite{Dolhansky2020}, DFDC Preview \cite{Dolhansky2019}, and DeeperForensics-1.0 \cite{jiang2020deeperforensics}. Experimental results demonstrate that the proposed approach is superior to state-of-the-art methods. It however may have a limitation when dealing with fake images that are generated by methods that directly output the whole images whose source features are consistent across all positions within each image.

\subsection{Fake Video Detection}
Most image detection methods cannot be used for videos because of the strong degradation of the frame data after video compression \cite{Afchar2018}. Furthermore, videos have temporal characteristics that are varied among sets of frames and they are thus challenging for methods designed to detect only still fake images. This subsection focuses on deepfake video detection methods and categorizes them into two smaller groups: methods that employ temporal features and those that explore visual artifacts within frames.

\subsubsection{Temporal Features across Video Frames}
Based on the observation that temporal coherence is not enforced effectively in the synthesis process of deepfakes, Sabir et al. \cite{Sabir2019} leveraged the use of spatio-temporal features of video streams to detect deepfakes. Video manipulation is carried out on a frame-by-frame basis so that low level artifacts produced by face manipulations are believed to further manifest themselves as temporal artifacts with inconsistencies across frames. A recurrent convolutional model (RCN) was proposed based on the integration of the convolutional network DenseNet \cite{Huang2017} and the gated recurrent unit cells \cite{Cho2014} to exploit temporal discrepancies across frames (see Fig. \ref{fig3}). The proposed method is tested on the FaceForensics++ dataset, which includes 1,000 videos \cite{Rossler2019}, and shows promising results. 

\begin{figure}[!ht]
\centering
\includegraphics[width=1\columnwidth]{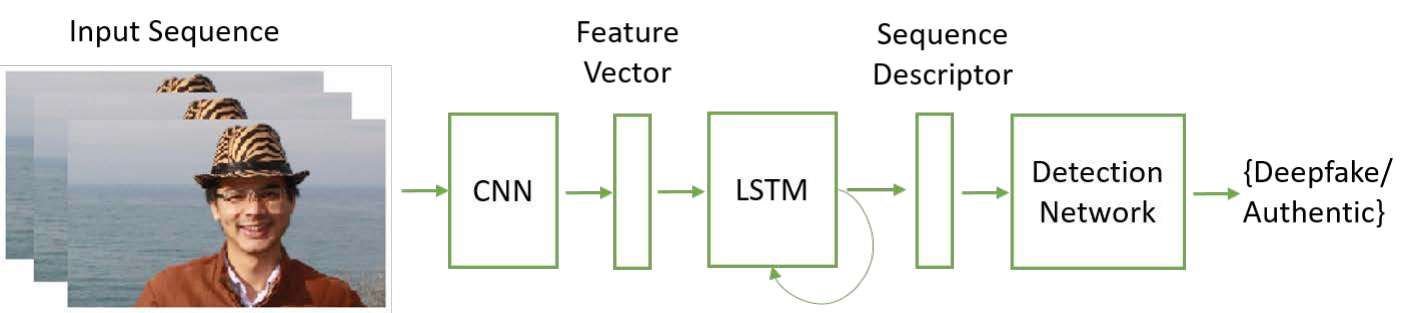}
\caption{A deepfake detection method using convolutional neural network (CNN) and long short term memory (LSTM) to extract temporal features of a given video sequence, which are represented via the sequence descriptor. The detection network consisting of fully-connected layers is employed to take the sequence descriptor as input and calculate probabilities of the frame sequence belonging to either authentic or deepfake class \cite{Guera2018}.}
\label{fig4} 
\end{figure}

Likewise, \citet{Guera2018} highlighted that deepfake videos contain intra-frame inconsistencies and temporal inconsistencies between frames. They then proposed the temporal-aware pipeline method that uses CNN and long short term memory (LSTM) to detect deepfake videos. CNN is employed to extract frame-level features, which are then fed into the LSTM to create a temporal sequence descriptor. A fully-connected network is finally used for classifying doctored videos from real ones based on the sequence descriptor as illustrated in Fig. \ref{fig4}. An accuracy of greater than 97\% was obtained using a dataset of 600 videos, including 300 deepfake videos collected from multiple video-hosting websites and 300 pristine videos randomly selected from the Hollywood human actions dataset in \cite{laptev2008learning}.

On the other hand, the use of a physiological signal, eye blinking, to detect deepfakes was proposed in Li et al. \cite{Li2018} based on the observation that a person in deepfakes has a lot less frequent blinking than that in untampered videos. A healthy adult human would normally blink somewhere between 2 to 10 seconds, and each blink would take 0.1 and 0.4 seconds. Deepfake algorithms, however, often use face images available online for training, which normally show people with open eyes, i.e., very few images published on the internet show people with closed eyes. Thus, without having access to images of people blinking, deepfake algorithms do not have the capability to generate fake faces that can blink normally. In other words, blinking rates in deepfakes are much lower than those in normal videos. To discriminate real and fake videos, Li et al. \cite{Li2018} crop eye areas in the videos and distribute them into long-term recurrent convolutional networks (LRCN) \cite{Donahue2015} for dynamic state prediction. The LRCN consists of a feature extractor based on CNN, a sequence learning based on long short term memory (LSTM), and a state prediction based on a fully connected layer to predict probability of eye open and close state. The eye blinking shows strong temporal dependencies and thus the implementation of LSTM helps to capture these temporal patterns effectively.

Recently, \citet{caldelli2021optical} proposed the use of optical flow to gauge the information along the temporal axis of a frame sequence for video deepfake detection. The optical flow is a vector field calculated on two temporal-distinct frames of a video that can describe the movement of objects in a scene. The optical flow fields are expected to be different between synthetically created frames and naturally generated ones \cite{amerini2019deepfake}. Unnatural movements of lips, eyes, or of the entire faces inserted into deepfake videos would introduce distinctive motion patterns when compared with pristine ones. Based on this assumption, features consisting of optical flow fields are fed into a CNN model for discriminating between deepfakes and original videos. More specifically, the ResNet50 architecture \cite{He2016} is implemented as a CNN model for experiments. The results obtained using the FaceForensics++ dataset \cite{Rossler2019} show that this approach is comparable with state-of-the-art methods in terms of classification accuracy. A combination of this kind of feature with frame-based features is also experimented, which results in an improved deepfake detection performance. This demonstrates the usefulness of optical flow fields in capturing the inconsistencies on the temporal axis of video frames for deepfake detection.

\subsubsection{Visual Artifacts within Video Frame}
As can be noticed in the previous subsection, the methods using temporal patterns across video frames are mostly based on deep recurrent network models to detect deepfake videos. This subsection investigates the other approach that normally decomposes videos into frames and explores visual artifacts within single frames to obtain discriminant features. These features are then distributed into either a deep or shallow classifier to differentiate between fake and authentic videos. We thus group methods in this subsection based on the types of classifiers, i.e. either deep or shallow.

\paragraph{Deep classifiers}
Deepfake videos are normally created with limited resolutions, which require an affine face warping approach (i.e., scaling, rotation and shearing) to match the configuration of the original ones. Because of the resolution inconsistency between the warped face area and the surrounding context, this process leaves artifacts that can be detected by CNN models such as VGG16 \cite{Simonyan2014}, ResNet50, ResNet101 and ResNet152 \cite{He2016}. A deep learning method to detect deepfakes based on the artifacts observed during the face warping step of the deepfake generation algorithms was proposed in \cite{Li2019}. The proposed method is evaluated on two deepfake datasets, namely the UADFV and DeepfakeTIMIT. The UADFV dataset \cite{Yang2019} contains 49 real videos and 49 fake videos with 32,752 frames in total. The DeepfakeTIMIT dataset \cite{Korshunov2018} includes a set of low quality videos of 64 x 64 size and another set of high quality videos of 128 x 128 with totally 10,537 pristine images and 34,023 fabricated images extracted from 320 videos for each quality set. Performance of the proposed method is compared with other prevalent methods such as two deepfake detection MesoNet methods, i.e. Meso-4 and MesoInception-4 \cite{Afchar2018}, HeadPose \cite{Yang2019}, and the face tampering detection method two-stream NN \cite{Zhou2015}. Advantage of the proposed method is that it needs not to generate deepfake videos as negative examples before training the detection models. Instead, the negative examples are generated dynamically by extracting the face region of the original image and aligning it into multiple scales before applying Gaussian blur to a scaled image of random pick and warping back to the original image. This reduces a large amount of time and computational resources compared to other methods, which require deepfakes are generated in advance.

\begin{figure}[!ht]
\centering
\includegraphics[width=1\columnwidth]{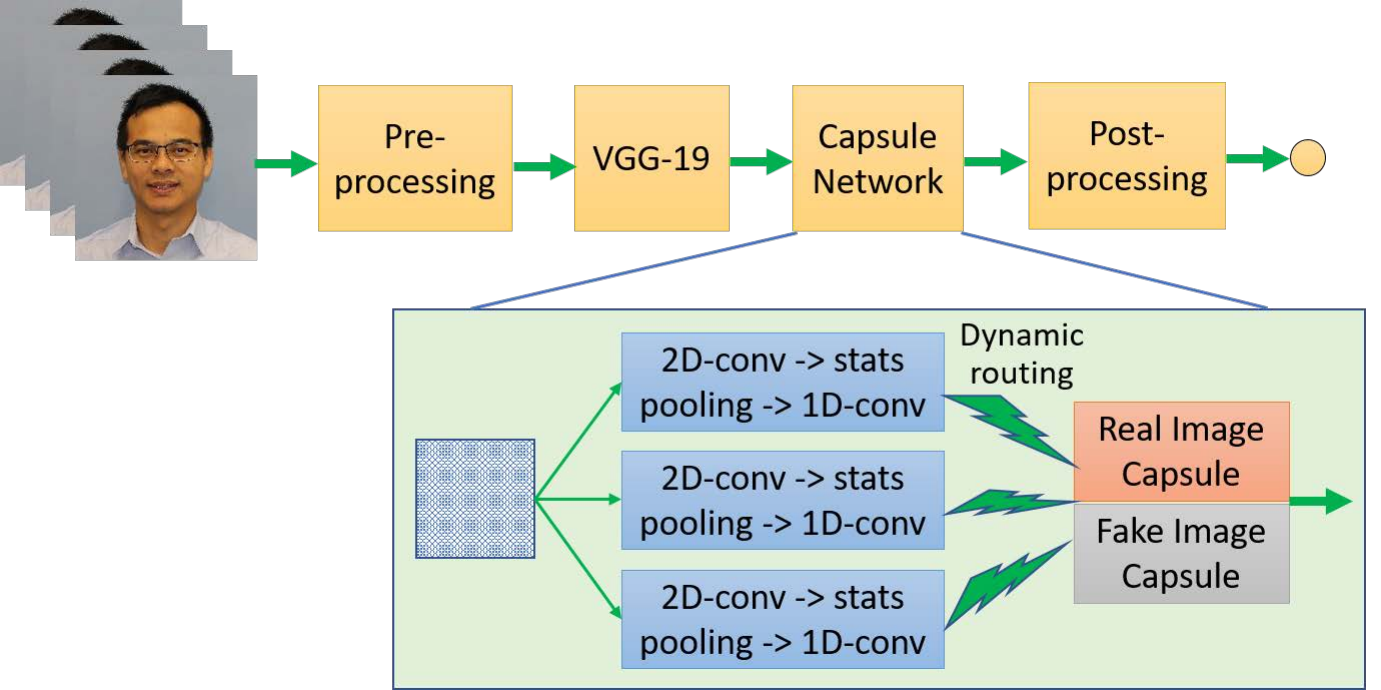}
\caption{Capsule network takes features obtained from the VGG-19 network \cite{Simonyan2014} to distinguish fake images or videos from the real ones (top). The pre-processing step detects face region and scales it to the size of 128x128 before VGG-19 is used to extract latent features for the capsule network, which comprises three primary capsules and two output capsules, one for real and one for fake images (bottom). The statistical pooling constitutes an important part of capsule network that deals with forgery detection \cite{Nguyen2019}.}
\label{fig5} 
\end{figure}

Nguyen et al. \cite{Nguyen2019} proposed the use of capsule networks for detecting manipulated images and videos. The capsule network was initially introduced to address limitations of CNNs when applied to inverse graphics tasks, which aim to find physical processes used to produce images of the world \cite{Hinton2011}. The recent development of capsule network based on dynamic routing algorithm \cite{Sabour2017} demonstrates its ability to describe the hierarchical pose relationships between object parts. This development is employed as a component in a pipeline for detecting fabricated images and videos as demonstrated in Fig. \ref{fig5}. A dynamic routing algorithm is deployed to route the outputs of the three capsules to the output capsules through a number of iterations to separate between fake and real images. The method is evaluated through four datasets covering a wide range of forged image and video attacks. They include the well-known Idiap Research Institute replay-attack dataset \cite{Chingovska2012}, the deepfake face swapping dataset created by Afchar et al. \cite{Afchar2018}, the facial reenactment FaceForensics dataset \cite{Rossler2018}, produced by the Face2Face method \cite{Thies2016}, and the fully computer-generated image dataset generated by Rahmouni et al. \cite{Rahmouni2017}. The proposed method yields the best performance compared to its competing methods in all of these datasets. This shows the potential of the capsule network in building a general detection system that can work effectively for various forged image and video attacks.

\paragraph{Shallow classifiers}
Deepfake detection methods mostly rely on the artifacts or inconsistency of intrinsic features between fake and real images or videos. Yang et al. \cite{Yang2019} proposed a detection method by observing the differences between 3D head poses comprising head orientation and position, which are estimated based on 68 facial landmarks of the central face region. The 3D head poses are examined because there is a shortcoming in the deepfake face generation pipeline. The extracted features are fed into an SVM classifier to obtain the detection results. Experiments on two datasets show the great performance of the proposed approach against its competing methods. The first dataset, namely UADFV, consists of 49 deep fake videos and their respective real videos \cite{Yang2019}. The second dataset comprises 241 real images and 252 deep fake images, which is a subset of data used in the DARPA MediFor GAN Image/Video Challenge \cite{Guan2019}. Likewise, a method to exploit artifacts of deepfakes and face manipulations based on visual features of eyes, teeth and facial contours was studied in \cite{Matern2019}. The visual artifacts arise from lacking global consistency, wrong or imprecise estimation of the incident illumination, or imprecise estimation of the underlying geometry. For deepfakes detection, missing reflections and missing details in the eye and teeth areas are exploited as well as texture features extracted from the facial region based on facial landmarks. Accordingly, the eye feature vector, teeth feature vector and features extracted from the full-face crop are used. After extracting the features, two classifiers including logistic regression and small neural network are employed to classify the deepfakes from real videos. Experiments carried out on a video dataset downloaded from YouTube show the best result of 0.851 in terms of the area under the receiver operating characteristics curve. The proposed method however has a disadvantage that requires images meeting certain prerequisite such as open eyes or visual teeth.

The use of photo response non uniformity (PRNU) analysis was proposed in \cite{Koopman2018} to detect deepfakes from authentic ones. PRNU is a component of sensor pattern noise, which is attributed to the manufacturing imperfection of silicon wafers and the inconsistent sensitivity of pixels to light because of the variation of the physical characteristics of the silicon wafers. The PRNU analysis is widely used in image forensics \cite{Rosenfeld2009, Li2012, Lin2017, Scherhag2019, Phan2019} and advocated to use in \cite{Koopman2018} because the swapped face is supposed to alter the local PRNU pattern in the facial area of video frames. The videos are converted into frames, which are cropped to the questioned facial region. The cropped frames are then separated sequentially into eight groups where an average PRNU pattern is computed for each group. Normalised cross correlation scores are calculated for comparisons of PRNU patterns among these groups. A test dataset was created, consisting of 10 authentic videos and 16 manipulated videos, where the fake videos were produced from the genuine ones by the DeepFaceLab tool \cite{DeepFaceLabweb}. The analysis shows a significant statistical difference in terms of mean normalised cross correlation scores between deepfakes and the genuine. This analysis therefore suggests that PRNU has a potential in deepfake detection although a larger dataset would need to be tested. 

When seeing a video or image with suspicion, users normally want to search for its origin. However, there is currently no feasibility for such a tool. Hasan and Salah \cite{Hasan2019} proposed the use of blockchain and smart contracts to help users detect deepfake videos based on the assumption that videos are only real when their sources are traceable. Each video is associated with a smart contract that links to its parent video and each parent video has a link to its child in a hierarchical structure. Through this chain, users can credibly trace back to the original smart contract associated with pristine video even if the video has been copied multiple times. An important attribute of the smart contract is the unique hashes of the interplanetary file system, which is used to store video and its metadata in a decentralized and content-addressable manner \cite{IPFS2019}. The smart contract's key features and functionalities are tested against several common security challenges such as distributed denial of services, replay and man in the middle attacks to ensure the solution meeting security requirements. This approach is generic, and it can be extended to other types of digital content, e.g., images, audios and manuscripts.

\begin{table*}[h!]
\centering
\caption{Summary of prominent deepfake detection methods}
\label{table2}
\begin{scriptsize}
\begin{tabularx}{\linewidth}{p{0.11\textwidth} p{0.09\textwidth} p{0.33\textwidth} p{0.04\textwidth} p{0.32\textwidth}}
\toprule
\textbf{Methods} & \textbf{Classifiers/\newline Techniques} & \textbf{Key Features} & \textbf{Dealing with} & \textbf{Datasets Used}\\
\midrule
%\begin{center}
%\begin{scriptsize}
%\begin{longtable}{p{0.11\textwidth} p{0.09\textwidth} p{0.32\textwidth} p{0.05\textwidth} p{0.32\textwidth}}
%\caption{SUMMARY OF PROMINENT DEEPFAKE DETECTION METHODS} \label{table2} \\
%\hline {\textbf{Methods}} & {\textbf{Classifiers/\newline Techniques}} & {\textbf{Key Features}}  & {\textbf{Dealing\newline with}}  & {\textbf{Datasets Used}} \\ \hline 
%\endfirsthead
%\multicolumn{5}{c}%
%{} \\
%\hline {\textbf{Methods}} & {\textbf{Classifiers/\newline Techniques}} & {\textbf{Key Features}}  & {\textbf{Dealing\newline with}}  & {\textbf{Datasets Used}} \\ \hline 
%\endhead
%\multicolumn{5}{r}{{Continued on next page}} \\
%\endfoot
%\hline
%\endlastfoot
Eye blinking \cite{Li2018}
&LRCN
&- Use LRCN to learn the temporal patterns of eye blinking.\newline
- Based on the observation that blinking frequency of deepfakes is much smaller than normal.
&Videos
&Consist of 49 interview and presentation videos, and their corresponding generated deepfakes.\\
\hline
Intra-frame and temporal inconsistencies \cite{Guera2018}
&CNN and LSTM
&CNN is employed to extract frame-level features, which are distributed to LSTM to construct sequence descriptor useful for classification.	
&Videos
&A collection of 600 videos obtained from multiple websites.\\
\hline
Using face warping artifacts \cite{Li2019}
&VGG16 \cite{Simonyan2014},\newline ResNet models \cite{He2016}
&Artifacts are discovered using CNN models based on resolution inconsistency between the warped face area and the surrounding context.
&Videos
&- UADFV \cite{Yang2019}, containing 49 real videos and 49 fake videos with 32752 frames in total.\newline
- DeepfakeTIMIT \cite{Korshunov2018}\\
\hline
MesoNet \cite{Afchar2018}
&CNN
&- Two deep networks, i.e. Meso-4 and MesoInception-4 are introduced to examine deepfake videos at the mesoscopic analysis level.\newline
- Accuracy obtained on deepfake and FaceForensics datasets are 98\% and 95\% respectively.
&Videos
&Two datasets: deepfake one constituted from online videos and the FaceForensics one created by the Face2Face approach \cite{Thies2016}.\\
\hline
Eye, teach and facial texture \cite{Matern2019}
&Logistic regression and neural network (NN)
&- Exploit facial texture differences, and missing reflections and details in eye and teeth areas of deepfakes.\newline
- Logistic regression and NN are used for classifying.	
&Videos
&A video dataset downloaded from YouTube.\\
\hline
Spatio-temporal features with RCN \cite{Sabir2019}
&RCN
&Temporal discrepancies across frames are explored using RCN that integrates convolutional network DenseNet \cite{Huang2017} and the gated recurrent unit cells \cite{Cho2014}
&Videos
&FaceForensics++ dataset, including 1,000 videos \cite{Rossler2019}.\\
\hline
Spatio-temporal features with LSTM \cite{Chintha2020}
& Convolutional bidirectional recurrent LSTM network
& - An XceptionNet CNN is used for facial feature extraction while audio embeddings are obtained by stacking multiple convolution modules.\newline
- Two loss functions, i.e. cross-entropy and Kullback-Leibler divergence, are used.
& Videos
& FaceForensics++ \cite{Rossler2019} and Celeb-DF (5,639 deepfake videos) \cite{Li2020} datasets and the ASVSpoof 2019 Logical Access audio dataset \cite{Todisco2019}.\\
\hline
Analysis of PRNU \cite{Koopman2018}
&PRNU
&- Analysis of noise patterns of light sensitive sensors of digital cameras due to their factory defects.\newline
- Explore the differences of PRNU patterns between the authentic and deepfake videos because face swapping is believed to alter the local PRNU patterns.
&Videos	
&Created by the authors, including 10 authentic and 16 deepfake videos using DeepFaceLab \cite{DeepFaceLabweb}.\\
\hline
Phoneme-viseme mismatches \cite{Agarwal2020a}
& CNN
& - Exploit the mismatches between the dynamics of the mouth shape, i.e. visemes, with a spoken phoneme.\newline
- Focus on sounds associated with the M, B and P phonemes as they require complete mouth closure while deepfakes often incorrectly synthesize it.
& Videos
& Four in-the-wild lip-sync deepfakes from Instagram and YouTube (www.instagram.com/bill\_posters\_uk and youtu.be/VWMEDacz3L4) and others are created using synthesis techniques, i.e. Audio-to-Video (A2V) \cite{Suwajanakorn2017} and Text-to-Video (T2V) \cite{Fried2019}.\\
\hline
Using attribution-based confidence (ABC) metric \cite{Fernandes2020}
& ResNet50 model \cite{He2016}, pre-trained on VGGFace2 \cite{Cao2018}
& - The ABC metric \cite{Jha2019} is used to detect deepfake videos without accessing to training data.\newline
- ABC values obtained for original videos are greater than 0.94 while those of deepfakes have low ABC values.
& Videos
& VidTIMIT and two other original datasets obtained from the COHFACE (https://www.idiap.ch/dataset/cohface) and from YouTube. datasets from COHFACE \cite{Fernandes2019c} and YouTube are used to generate two deepfake datasets by commercial website https://deepfakesweb.com and another deepfake dataset is DeepfakeTIMIT \cite{Korshunov2018c}.\\
\hline
Using appearance and behaviour \cite{Agarwal2020e}
& Rules based on facial and behavioural features.
& Temporal, behavioral biometric based on facial expressions and head movements are learned using ResNet-101 \cite{He2016} while static facial biometric is obtained using VGG \cite{Parkhi2015}.
& Videos
& The world leaders dataset \cite{Agarwal2019d}, FaceForensics++ \cite{Rossler2019}, Google/Jigsaw deepfake detection dataset \cite{Dufour2019}, DFDC \cite{Dolhansky2019} and Celeb-DF \cite{Li2020}.\\
\hline
FakeCatcher \cite{Ciftci2020}
& CNN
& Extract biological signals in portrait videos and use them as an implicit descriptor of authenticity because they are not spatially and temporally well-preserved in deepfakes.
& Videos
& UADFV \cite{Yang2019}, FaceForensics \cite{Rossler2018}, FaceForensics++ \cite{Rossler2019}, Celeb-DF \cite{Li2020}, and a new dataset of 142 videos, independent of the generative model, resolution, compression, content, and context.\\
\hline
Emotion audio-visual affective cues \cite{Mittal2020}
& Siamese network \cite{Chopra2005}
& Modality and emotion embedding vectors for the face and speech are extracted for deepfake detection.
& Videos
& DeepfakeTIMIT \cite{Korshunov2018c} and DFDC \cite{Dolhansky2019}.\\
\hline
Head poses \cite{Yang2019}
&SVM
&- Features are extracted using 68 landmarks of the face region.\newline
- Use SVM to classify using the extracted features.	
&Videos/\newline Images
&- UADFV consists of 49 deep fake videos and their respective real videos.\newline
- 241 real images and 252 deep fake images from DARPA MediFor GAN Image/Video Challenge.\\
\hline
Capsule-forensics \cite{Nguyen2019}
&Capsule networks
&- Latent features extracted by VGG-19 network \cite{Simonyan2014} are fed into the capsule network for classification.\newline
- A dynamic routing algorithm \cite{Sabour2017} is used to route the outputs of three convolutional capsules to two output capsules, one for fake and another for real images, through a number of iterations.	
&Videos/\newline Images
&Four datasets: the Idiap Research Institute replay-attack \cite{Chingovska2012}, deepfake face swapping by \cite{Afchar2018}, facial reenactment FaceForensics \cite{Rossler2018}, and fully computer-generated image set using \cite{Rahmouni2017}.\\
\hline
\end{tabularx}
\end{scriptsize}
\end{table*}
\begin{table*}[h!]
\centering
%\ContinuedFloat
\begin{scriptsize}
\begin{tabularx}{\linewidth}{p{0.1\textwidth} p{0.1\textwidth} p{0.33\textwidth} p{0.04\textwidth} p{0.32\textwidth}}
\toprule
\textbf{Methods} & \textbf{Classifiers/\newline Techniques} & \textbf{Key Features} & \textbf{Dealing with} & \textbf{Datasets Used}\\
\midrule
Preprocessing combined with deep network \cite{Xuan2019}
&DCGAN,
WGAN-GP and PGGAN.
&- Enhance generalization ability of deep learning models to detect GAN generated images.\newline
- Remove low level features of fake images.\newline
- Force deep networks to focus more on pixel level similarity between fake and real images to improve generalization ability.
&Images
&- Real dataset: CelebA-HQ \cite{Karras2017}, including high quality face images of 1024x1024 resolution.\newline
- Fake datasets: generated by DCGAN \cite{Radford2015}, WGAN-GP \cite{Gulrajani2017} and PGGAN \cite{Karras2017}.\\
\hline
Analyzing convolutional traces \cite{Guarnera2020}
& KNN, SVM, and linear discriminant analysis (LDA)
& Using expectation-maximization algorithm to extract local features pertaining to convolutional generative process of GAN-based image deepfake generators.
& Images
& Authentic images from CelebA and corresponding deepfakes are created by five different GANs (group-wise deep whitening-and-coloring transformation GDWCT \cite{Cho2019}, StarGAN \cite{Choi2018},
AttGAN \cite{He2019}, StyleGAN \cite{Karras2019}, StyleGAN2 \cite{Karras2020}).\\
\hline
Bag of words and shallow classifiers \cite{Zhang2017}	
&SVM, RF, MLP
&Extract discriminant features using bag of words method and feed these features into SVM, RF and MLP for binary classification: innocent vs fabricated.
&Images
&The well-known LFW face database \cite{Huang2007}, containing 13,223 images with resolution of 250x250.\\
\hline
Pairwise learning \cite{Hsu2019}
&CNN concatenated to CFFN
&Two-phase procedure: feature extraction using CFFN based on the Siamese network architecture \cite{Chopra2005} and classification using CNN.
&Images
&- Face images: real ones from CelebA \cite{Liu2015}, and fake ones generated by DCGAN \cite{Radford2015}, WGAN \cite{Arjovsky2017}, WGAN-GP \cite{Gulrajani2017}, least squares GAN \cite{Mao2017}, and PGGAN \cite{Karras2017}.\newline
- General images: real ones from ILSVRC12 \cite{Russakovsky2015}, and fake ones generated by BIGGAN \cite{Brock2018}, self-attention GAN \cite{Zhang2018} and spectral normalization GAN \cite{Miyato2018}.\\
\hline
Defenses against adversarial perturbations in deepfakes \cite{Gandhi2020}
& VGG \cite{Parkhi2015} and ResNet \cite{He2016}
& - Introduce adversarial perturbations to enhance deepfakes and fool deepfake detectors.\newline
- Improve accuracy of deepfake detectors using Lipschitz regularization and deep image prior techniques.
& Images
& 5,000 real images from CelebA \cite{Liu2015} and 5,000 fake images created by the “Few-Shot Face Translation GAN” method \cite{Shaoanlu2019}.\\
\hline
Face X-ray \cite{Li2020e}
& CNN
& - Try to locate the blending boundary between the target and original faces instead of capturing the synthesized artifacts of specific manipulations.\newline
- Can be trained without fake images.
& Images
& FaceForensics++ \cite{Rossler2019}, DeepfakeDetection (DFD) \cite{Dufour2019}, DFDC \cite{Dolhansky2019} and Celeb-DF \cite{Li2020}.\\
\hline
Using common artifacts of CNN-generated images \cite{Wang2020}
& ResNet-50 \cite{He2016}
pre-trained with ImageNet \cite{Russakovsky2015}
& Train the classifier using a large number of fake images generated by a high-performing unconditional GAN model, i.e., PGGAN \cite{Karras2017} and evaluate how well the classifier generalizes to other CNN-synthesized images.
& Images
& A new dataset of CNN-generated images, namely ForenSynths, consisting of
synthesized images from 11 models such as StyleGAN \cite{Karras2019}, super-resolution methods \cite{Dai2019} and FaceForensics++ \cite{Rossler2019}.\\
\hline
Using convolutional traces on GAN-based images \cite{guarnera2020fighting}
& KNN, SVM, and LDA
& Training the expectation-maximization algorithm \cite{moon1996expectation} to detect
and extract discriminative features via a fingerprint that represents the convolutional traces left by GANs during image generation.
& Images
& A dataset of images generated by ten GAN models, including CycleGAN \cite{zhu2017unpaired}, StarGAN \cite{Choi2018}, AttGAN \cite{He2019}, GDWCT \cite{Cho2019}, StyleGAN \cite{Karras2019}, StyleGAN2 \cite{Karras2020}, PGGAN \cite{Karras2017}, FaceForensics++ \cite{Rossler2019}, IMLE \cite{li2019diverse}, and
SPADE \cite{Park2019e}.\\
\hline
Using deep features extracted by CNN \cite{guo2021blind}
& A new CNN model, namely SCnet
& The CNN-based SCnet is able to automatically learn high-level forensics features of image data thanks to a hierarchical feature extraction block, which is formed by stacking four convolutional layers.
& Images
& A dataset of 321,378 face images, created by applying the Glow model \cite{kingma2018glow} to the CelebA face image dataset \cite{Liu2015}.
\\

%\end{longtable}
%\end{scriptsize}
%\end{center}
\bottomrule
\end{tabularx}
\end{scriptsize}
\end{table*}

\section{Discussions and Future Research Directions}
\label{sec:4}
%disruptive technology.
%Informative graphics are provided for guiding readers through the latest development in deepfake research. 
With the support of deep learning, deepfakes can be created easier than ever before. The spread of these fake contents is also quicker thanks to the development of social media platforms \cite{Zubiaga2018}. Sometimes deepfakes do not need to be spread to massive audience to cause detrimental effects. People who create deepfakes with malicious purpose only need to deliver them to target audiences as part of their sabotage strategy without using social media. For example, this approach can be utilized by intelligence services trying to influence decisions made by important people such as politicians, leading to national and international security threats \cite{cfr}. Catching the deepfake alarming problem, research community has focused on developing deepfake detection algorithms and numerous results have been reported. This paper has reviewed the state-of-the-art methods and a summary of typical approaches is provided in Table \ref{table2}. It is noticeable that a battle between those who use advanced machine learning to create deepfakes with those who make effort to detect deepfakes is growing.

Deepfakes' quality has been increasing and the performance of detection methods needs to be improved accordingly. The inspiration is that what AI has broken can be fixed by AI as well \cite{Floridi2018}. Detection methods are still in their early stage and various methods have been proposed and evaluated but using fragmented datasets. An approach to improve performance of detection methods is to create a growing updated benchmark dataset of deepfakes to validate the ongoing development of detection methods. This will facilitate the training process of detection models, especially those based on deep learning, which requires a large training set \cite{Dolhansky2020}. 

Improving performance of deepfake detection methods is important, especially in cross-forgery and cross-dataset scenarios. Most detection models are designed and evaluated in the same-forgery and in-dataset experiments, which do not ensure their generalization capability. Some previous studies have addressed this issue, e.g., in \cite{Wang2020, caldelli2021optical, zhao2021learning, cozzolino2018forensictransfer, marra2019incremental}, but more work needs to be done in this direction. A model trained on a specific forgery needs to be able to work against another unknown one because potential deepfake types are not normally known in the real-world scenarios. Likewise, current detection methods mostly focus on drawbacks of the deepfake generation pipelines, i.e., finding weakness of the competitors to attack them. This kind of information and knowledge is not always available in adversarial environments where attackers commonly attempt not to reveal such deepfake creation technologies. Recent works on adversarial perturbation attacks to fool DNN-based detectors make the deepfake detection task more difficult \cite{Gandhi2020, Neekhara2020, Carlini2020, Yang2020, Yeh2020}. These are real challenges for detection method development and a future study needs to focus on introducing more robust, scalable and generalizable methods.

Another research direction is to integrate detection methods into distribution platforms such as social media to increase its effectiveness in dealing with the widespread impact of deepfakes. The screening or filtering mechanism using effective detection methods can be implemented on these platforms to ease the deepfakes detection \cite{cfr}. Legal requirements can be made for tech companies who own these platforms to remove deepfakes quickly to reduce its impacts. In addition, watermarking tools can also be integrated into devices that people use to make digital contents to create immutable metadata for storing originality details such as time and location of multimedia contents as well as their untampered attestment \cite{cfr}. This integration is difficult to implement but a solution for this could be the use of the disruptive blockchain technology. The blockchain has been used effectively in many areas and there are very few studies so far addressing the deepfake detection problems based on this technology. As it can create a chain of unique unchangeable blocks of metadata, it is a great tool for digital provenance solution. The integration of blockchain technologies to this problem has demonstrated certain results \cite{Hasan2019} but this research direction is far from mature.

Using detection methods to spot deepfakes is crucial, but understanding the real intent of people publishing deepfakes is even more important. This requires the judgement of users based on social context in which deepfake is discovered, e.g. who distributed it and what they said about it \cite{Read2019web}. This is critical as deepfakes are getting more and more photorealistic and it is highly anticipated that detection software will be lagging behind deepfake creation technology. A study on social context of deepfakes to assist users in such judgement is thus worth performing. 

Videos and photographics have been widely used as evidences in police investigation and justice cases. They may be introduced as evidences in a court of law by digital media forensics experts who have background in computer or law enforcement and experience in collecting, examining and analysing digital information. The development of machine learning and AI technologies might have been used to modify these digital contents and thus the experts' opinions may not be enough to authenticate these evidences because even experts are unable to discern manipulated contents. This aspect needs to take into account in courtrooms nowadays when images and videos are used as evidences to convict perpetrators because of the existence of a wide range of digital manipulation methods \cite{Maras2019}. The digital media forensics results therefore must be proved to be valid and reliable before they can be used in courts. This requires careful documentation for each step of the forensics process and how the results are reached. Machine learning and AI algorithms can be used to support the determination of the authenticity of digital media and have obtained accurate and reliable results, e.g., \cite{Su2017, Iuliani2018}, but most of these algorithms are unexplainable. This creates a huge hurdle for the applications of AI in forensics problems because not only the forensics experts oftentimes do not have expertise in computer algorithms, but the computer professionals also cannot explain the results properly as most of these algorithms are black box models \cite{Malolan2020}. This is more critical as the most recent models with the most accurate results are based on deep learning methods consisting of many neural network parameters. Researchers have recently attempted to create white box and explainable detection methods. An example is the approach proposed by \citet{giudice2021fighting} in which they use discrete cosine transform statistics to detect so-called specific GAN frequencies to differentiate between real images and deepfakes. Through the analysis of particular frequency statistics, that method can be used to mathematically explain whether a multimedia content is a deepfake and why it is. More research must be conducted in this area and explainable AI in computer vision therefore is a research direction that is needed to promote and utilize the advances and advantages of AI and machine learning in digital media forensics.

\section{Conclusions}
Deepfakes have begun to erode trust of people in media contents as seeing them is no longer commensurate with believing in them. They could cause distress and negative effects to those targeted, heighten disinformation and hate speech, and even could stimulate political tension, inflame the public, violence or war. This is especially critical nowadays as the technologies for creating deepfakes are increasingly approachable and social media platforms can spread those fake contents quickly. This survey provides a timely overview of deepfake creation and detection methods and presents a broad discussion on challenges, potential trends, and future directions in this area. This study therefore will be valuable for the artificial intelligence research community to develop effective methods for tackling deepfakes.

\section*{Declaration of Competing Interest}
Authors declare no conflict of interest.

% \bibliographystyle{elsarticle-num} 
%\bibliographystyle{plainnat}

\bibliography{cas-refs}





% \end{thebibliography}
\end{document}